\documentclass[lettersize,journal]{IEEEtran}
\usepackage{amsmath,amsfonts}
\usepackage{algorithmic}
\usepackage{algorithm}
\usepackage{array}
\usepackage[caption=false,font=normalsize,labelfont=sf,textfont=sf]{subfig}
\usepackage{textcomp}
\usepackage{stfloats}
\usepackage{url}
\usepackage{verbatim}
\usepackage{graphicx}
\usepackage{cite}

\usepackage{threeparttable}
\usepackage{tabularx}
\usepackage{booktabs}
\usepackage{multirow}
\usepackage{subfig}
\usepackage{color}
\usepackage{gensymb}

\newcolumntype{L}[1]{>{\raggedright\arraybackslash}p{#1}}
\newcolumntype{C}[1]{>{\centering\arraybackslash}p{#1}}
\newcolumntype{R}[1]{>{\raggedleft\arraybackslash}p{#1}}

\usepackage{tikz,xcolor,hyperref}
\definecolor{lime}{HTML}{A6CE39}
\DeclareRobustCommand{\orcidicon}{%
    \begin{tikzpicture}
    \draw[lime, fill=lime] (0,0) 
    circle [radius=0.16] 
    node[white] {{\fontfamily{qag}\selectfont \tiny ID}};    \draw[white, fill=white] (-0.0625,0.095) 
    circle [radius=0.007];    \end{tikzpicture}
    \hspace{-2mm}}
\foreach \x in {JLW, XJL, HYC, HD, YHY, ZH}{%
    \expandafter\xdef\csname orcid\x\endcsname{\noexpand\href{https://orcid.org/\csname orcidauthor\x\endcsname}{\noexpand\orcidicon}}
    }

\begin{document}

\title{MSN: Multi-directional Similarity Network for Hand-crafted and Deep-synthesized Copy-Move Forgery Detection}

\newcommand{\orcidauthorJLW}{0000-0002-8264-287X}
\newcommand{\orcidauthorXJL}{0000-0002-3662-4697}
\newcommand{\orcidauthorHYC}{0000-0002-5754-0361}
\newcommand{\orcidauthorHD}{0000-0002-2412-9330}
\newcommand{\orcidauthorYHY}{0000-0001-5763-1629}
\newcommand{\orcidauthorZH}{0000-0002-7627-4142}

\author{Liangwei Jiang\orcidJLW{}, Jinluo Xie\orcidXJL{}, Yecheng Huang\orcidHYC{}, Hua Zhang\orcidZH{}, Hongyu Yang\orcidYHY{}, \IEEEmembership{Member,~IEEE}, Di Huang\orcidHD{}, ~\IEEEmembership {Senior Member,~IEEE}
\thanks{
L. Jiang, J. Xie, Y. Huang, and D. Huang are with the State Key Laboratory of Software Development Environment, Beihang University, Beijing
100191, China. E-mail: {lw\_jiang, xiejinluo, ychuang, dhuang}@buaa.edu.cn.} 

\thanks{
H. Yang is with the Institute of Artificial Intelligence, Beihang University, Beijing
100191, China. E-mail: hongyuyang@buaa.edu.cn. (Corresponding author: Hongyu Yang.)
}
\thanks{
H. Zhang is with the State Key Laboratory of Information Security, Institute of Information Engineering, Chinese
Academy of Sciences, Beijing 100093, China. E-mail: zhanghua@iie.ac.cn.
}
\thanks{Manuscript received August 10, 2022.}

}



\maketitle

\begin{abstract}
Copy-move image forgery aims to duplicate certain objects or to hide specific contents with copy-move operations, which can be achieved by a sequence of manual manipulations as well as up-to-date deep generative network-based swapping. Its detection is becoming increasingly challenging for the complex transformations and fine-tuned operations on the tampered regions. In this paper, we propose a novel two-stream model, namely Multi-directional Similarity Network (MSN), to accurate and efficient copy-move forgery detection. It addresses the two major limitations of existing deep detection models in \textbf{representation} and \textbf{localization}, respectively. In representation, an image is hierarchically encoded by a multi-directional CNN network, and due to the diverse augmentation in scales and rotations, the feature achieved better measures the similarity between sampled patches in two streams. In localization, we design a 2-D similarity matrix based decoder, and compared with the current 1-D similarity vector based one, it makes full use of spatial information in the entire image, leading to the improvement in detecting tampered regions. Beyond the method, a new forgery database generated by various deep neural networks is presented, as a new benchmark for detecting the growing deep-synthesized copy-move. Extensive experiments are conducted on two classic image forensics benchmarks, \emph{i.e.} CASIA CMFD and CoMoFoD, and the newly presented one. The state-of-the-art results are reported, which demonstrate the effectiveness of the proposed approach.
\end{abstract}


\begin{figure}[tp]
    \flushright
	\includegraphics[width=1.0\columnwidth]{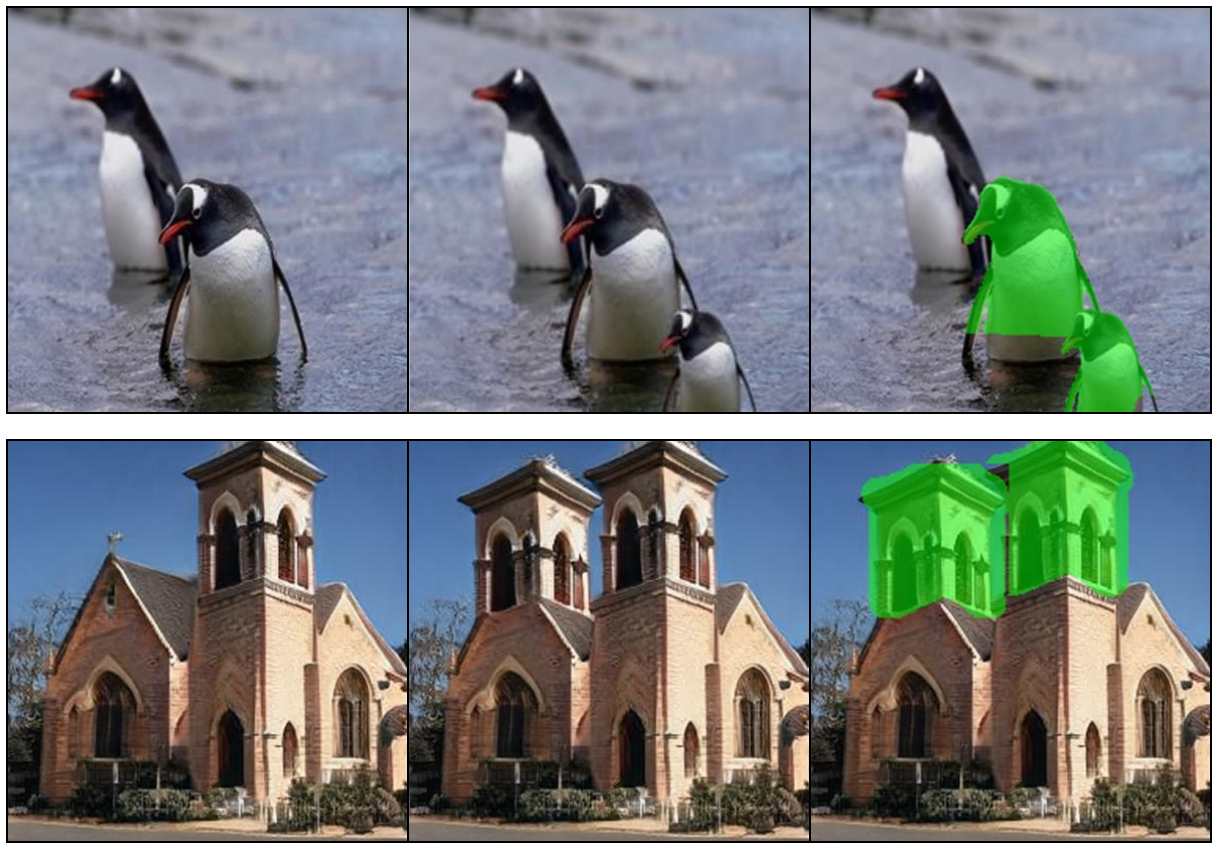}
	\caption{Two copy-move forgery examples achieved by manual manipulation (upper row) and deep generative network (bottom row). From left to right: original images, forged images, and tampered regions. }
	\label{cmfd example}
\end{figure}

\section{Introduction}
    Images convey abundant information in a direct and vivid manner. In recent years, the amount of images has been consistently and sharply increasing and images are almost everywhere. Meanwhile, image editing techniques have been rapidly developed and have substantially improved productivity in the multimedia and publishing industries. Unfortunately, they also provide convenience to image abuse and forgery, which probably interfere our understanding and incur negative impacts. For instance, in social apps, fake news and rumors usually use tampered images to make their stories more credible, while in academic papers, fraudulent research often fabricates experimental charts on previously published ones.
    
    Copy-move forgery is one of the most common and typical image manipulations, where a sub-region (referred to as \textbf{source region}) of an image is copied, transformed and pasted elsewhere (referred to as \textbf{target region}) within the same image, to hide or duplicate objects. This process is mostly achieved by a sequence of manual operations and it has lately been automated by up-to-date techniques such as Generative Adversarial Network-based (GANs) swapping \cite{goodfellow2014generative, HongF2GAN,  li2018beautygan, bau2020rewriting, zhu2017unpaired} (as depicted in Fig. \ref{cmfd example}). Copy-Move Forgery Detection (CMFD) aims to determine whether the given image undergoes such forgery and if it does, locating the relevant regions. Because the transformation of the source region is unknown in advance and often involves complex variations, such as rotation, compression, and noise adding, CMFD is a rather challenging issue.
    
    \IEEEpubidadjcol
    
    During the last two decades, many methods have been proposed in the literature \cite{DBLP:journals/mta/ZhuSC16,li2018fast,DBLP:conf/eccv/WuAN18, shivakumar2011detection,amerini2013copy,manu2016detection,amerini2013copy,yang2018copy}. The overwhelming majority generally conduct hand-crafted local feature based analysis and share a universal pipeline, consisting of three main phases, \emph{i.e.}, region representation, region matching, and post-processing. Regions are first sampled from the image, then projected into the feature space to highlight texture properties, and further compared based on matching similarities. Through refinement by certain constraints (\emph{e.g.}, removing false alarms through spatial relationship), the tampered positions are finally localized. Even though these methods report competitive results, they suffer from very low efficiency and thus are problematic in real-world applications. 
    
    The advent and success of deep learning techniques show a promising alternative to accurate and efficient CMFD \cite{DBLP:conf/eccv/WuAN18,DBLP:journals/tifs/ZhongP20}, benefitting from a hierarchical representation of Convolutional Neural Networks (CNNs) as well as parallel computation of Graphics Processing Unit (GPU). Compared to traditional methods, these attempts promote efficiency by an order of magnitude. Nevertheless, there still exist drawbacks. 
    On the one hand, CNNs are not inherently invariant to rotation and scaling, making the similarity between the source and target areas across such changes unreliable. Despite the use of some reputed modules to enhance this robustness \cite{zhu2020ar,DBLP:conf/cvpr/IslamLBH20, DBLP:journals/tifs/ZhongP20}, the results are not as good as expected, since rotation and scaling are more serious in CMFD than in general object detection.
    On the other hand, in the encoder-decoder models, the decoding mechanism is not strong enough to stably locate tampered regions. It only works on the sorted and sampled similarities of CNN features from image regions without considering the entire structure, thus limiting the accuracy. 
    
    Additionally, it is important to note that the emergence of deep detection methods are accompanied by a rapid evolution of the techniques for manipulating images \cite{HongF2GAN, li2018beautygan, lyu2021sogan, bau2020rewriting, ling2021editgan, wang2021sketch}. A number of neural network approaches \cite{goodfellow2014generative, kingma2013auto, Karras2019stylegan2, karras2017progressive} have shown the potential in generating images with visually appealing effects. These forged images may no longer convey clues that used to be exploited as the foundation for detection and localization, \textit{e.g.}, edge and color inconsistencies. Regrettably, to the best of our knowledge, how existing deep detection models are threatened by such tampering types has not been verified due to the lack of benchmarks.
    
    In this paper, we first propose a novel deep learning approach, namely Multi-directional Similarity Network (MSN), to accurate and efficient CMFD, which handles the shortcomings of current deep models in feature learning and region localization. Specifically, for the former, MSN builds a CNN with a multi-directional architecture for hierarchical image encoding. Due to the augmentation in scale and direction, this representation is more powerful and hence better measures the similarity between the source and target regions. For the latter, different from the 1-D similarity vector \cite{DBLP:conf/eccv/WuAN18,DBLP:journals/tifs/ZhongP20,DBLP:conf/cvpr/IslamLBH20}, a 2-D similarity matrix is computed on a set of similarity maps. An additional classifier is designed to judge if the region corresponding to the similarity map is tampered, which takes similarities with original structure cues preserved and spatial context aggregated, thereby achieving precision gain in decoding. 
    
    Furthermore, to investigate the detection performance on deep synthetic data, we create the first copy-move forgery database with samples generated by three different neural networks, which is complementary to the current hand-crafted ones for CMFD. We carry out comprehensive experiments on two major public benchmarks, \emph{i.e.} CASIA CMFD and CoMoFoD, and the newly presented one. The results achieved clearly validate the effectiveness of the proposed approach on manually manipulated data, while simultaneously revealing that almost all current detection methods encounter performance degradation of varying degrees on deep forgery data where our approach performs more favorably.
    
    More concisely, this study makes the following contributions:
    \begin{enumerate}
        \item  A novel learning based method for CMFD, which incorporates a multi-directional network coping with rotation and scaling variants, and a 2-D similarity matrix for improved tampered region segmentation, thereby addressing the issues of feature representation and localization more effectively;
        \item  A new benchmark for CMFD, which is forged by three types of deep generative models, providing deep synthetic copy-move data for more comprehensive evaluations of detection models, along with the baseline performance for fair comparison in future studies; 
        \item  Extensive evaluations, including the experiments on two major public datasets, as well as on the newly presented deep synthesized benchmark, with state-of-the-art results reported.
    \end{enumerate}
    The rest of this paper is organized as follows. Section \ref{related_work} reviews related work on copy-move forgery detection. Section \ref{Methodology} details the proposed Multi-directional Similarity Network. Section \ref{Experiments} displays and analyzes the experimental results on three databases, followed by Section \ref{conclusion} concluding this paper with perspectives.

	\begin{figure*}[ht]
	\centering
	\includegraphics[width=2.06\columnwidth]{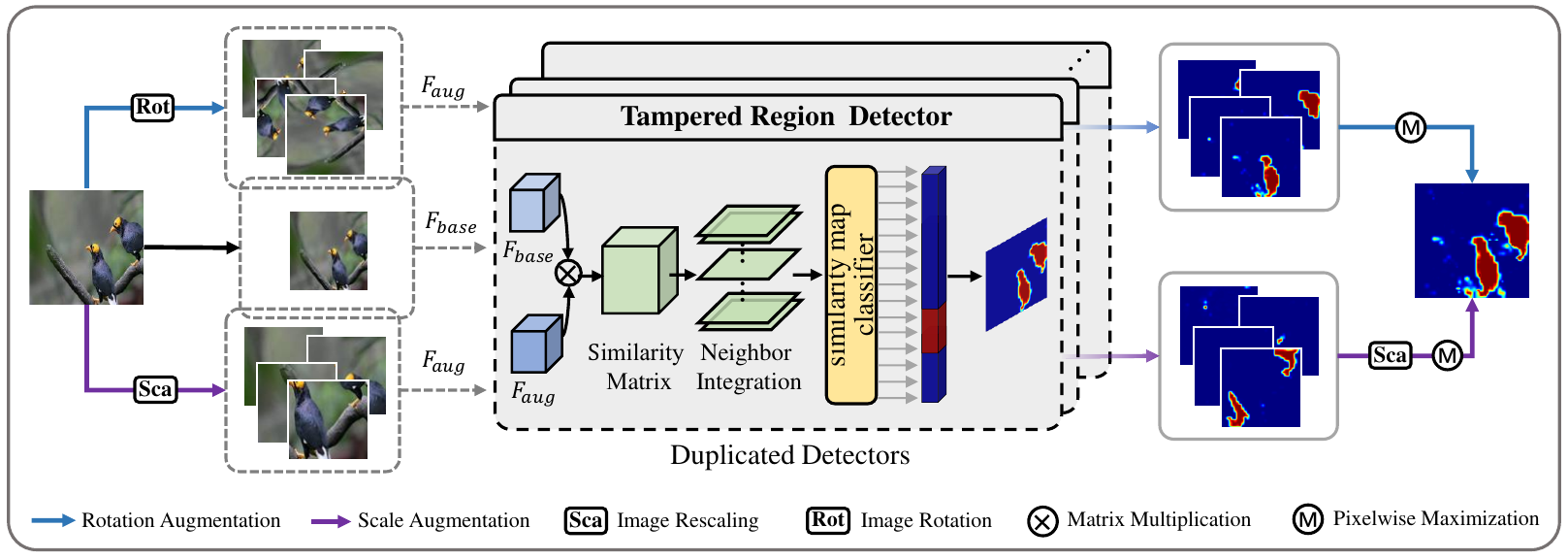}
	\caption{Framework overview. Given a query image, a multi-directional image set and a zoomed-in patch set are constructed by rotating it in four pre-defined quantized orientations and slicing out four patches from the enlarged image, respectively. The image or patch features are extracted through CNNs with shared weights, and then eight feature pairs are developed by fusing features from the basic input image and the augmented ones. They are individually fed into the two-stream detector to predict copy-move tampered region candidates, where a 2-D similarity matrix based decoder is designed for more accurate localization. The outputs of the eight duplicated detectors are adopted to jointly render a mask to localize the final tampered regions.
	Best viewed in color.}
	\label{overview}
	\end{figure*}

\section{Related Work}
\label{related_work}

	In the past decades, a large number of methods have been proposed to detect copy-move image forgeries. They basically follow the same framework of region representation, region matching, and post-processing, and their difference mainly lies in region representation in terms of region sampling and feature extraction. Most methods sparsely sample regions from the given images by keypoint detectors and local features are computed on them. In this stage, SIFT \cite{amerini2013copy,DBLP:journals/tifs/CostanzoACB14,yang2018copy}, ORB \cite{DBLP:journals/mta/ZhuSC16}, triangles \cite{DBLP:journals/tifs/ArdizzoneBM15}, and SURF \cite{shivakumar2011detection,DBLP:journals/jvcir/SilvaCFR15,manu2016detection} are widely exploited. Some efforts are made to divide images into irregular segments for representation to better fit the diversity of tampered regions \cite{DBLP:journals/tifs/LiLYS15,DBLP:journals/tifs/PunYB15}. Meanwhile, a number of methods adopt sliding windows to densely sample regions, where chroma features \cite{DBLP:conf/icassp/BayramSM09,DBLP:journals/tifs/CozzolinoPV15}, PCA coefficients \cite{DBLP:journals/mta/HuangHHC17}, Zernike moments \cite{DBLP:conf/ih/RyuLL10}, and DCT \cite{DBLP:journals/FEI/Mahmod17} prove more discriminative in this manner. Despite the high scores achieved on public datasets, all these methods are not qualified in more complex and arbitrary variations due to hand-crafted features. Furthermore, they require a rather long running time, making them impractical for applications.

	Recently, some investigations introduce Deep Neural Networks (DNNs) into CMFD and show promising results. \cite{DBLP:conf/eccv/WuAN18} presents the first  model which calculates similarities between CNN features and employs a decoder, where percentile pooling is applied to each sorted similarity vector to build statistics, to identify tampered and pristine regions. \cite{DBLP:conf/cvpr/IslamLBH20} proposes a Generative Adversarial Network (GAN) with a dual-order attention module to localize forgeries. In the generator, the attention module is designed to roughly locate all similar regions, where the 1st-order attention captures copy-move location clues whilst the 2nd-order attention models patch co-occurrence. Both attention maps are extracted from the affinity matrix and combined to feed into the detection and localization branches. The discriminator ensures more accurate localization predictions. \cite{DBLP:journals/tifs/ZhongP20} extracts multi-scale features and fuse feature correlation matching to deliver detection results. To localize tampered regions, a feature correlation matching module is built by predicting the regions which are abnormally closer to the nearest neighbor than the second nearest. But due to the incompetency of CNNs in dealing with rotation and scaling as well as the naive 1-D similarity based decoding, existing deep models leave much space for performance improvement.

	Contemporaneously, techniques for image generating and editing using neural networks have rapidly matured, especially the ones based on Generative Adversarial Networks (GANs) \cite{goodfellow2014generative, HongF2GAN, li2018beautygan}. For instance, the development of Progressive GAN \cite{karras2017progressive}, and StyleGAN2 \cite{Karras2019stylegan2} make it possible to generate very realistic and high-resolution images. CycleGAN \cite{zhu2017unpaired} and SOGAN \cite{lyu2021sogan} perform amazing image styling. Sketch-GAN \cite{wang2021sketch}, EditGAN \cite{ling2021editgan}, and Rewriting \cite{bau2020rewriting} are successfully applied for conditionalized image editing. However, there is a lack of copy-move forgery benchmark generated by these methods, and a certain gap exists between CMFD methods and generation techniques in the era of deep learning.
	
	In this study, we propose a novel and effective two-stream model to copy-move forgery detection. It addresses the two major limitations of existing deep detection models in representation and localization, respectively. Meanwhile, a new forgery benchmark generated by various deep neural networks is presented for more comprehensive evaluations.    
	
\section{Methodology}
\label{Methodology}
	
	{The entire framework of the proposed Multi-directional Similarity Network is illustrated in Fig. \ref{overview}. Given a query image, we first extend the input space by rotating it to predefined quantized orientations and rescaling it and slicing out the zoomed-in patches. Features are extracted from the augmented images by a CNN-based network. Eight feature pairs are then developed by fusing the feature representations from the original input image and the augmented ones. Thanks to the augmentation in scale and direction, the new multi-directional and multi-scale representation delivers more informative details and is more capable of handling the variations above. 
	
	In order to localize the tampered region candidates, each of the pairwise features is fed into a two-stream detector at the next stage, which consists of a Similarity Computation Layer and a Similarity Map Classifier. The former is introduced to measure the similarities between the sampled patches of the given image, while the latter is designed to decode the similarity matrix and predict the source and target regions. All the candidates are combined to generate the final mask of the tampered regions jointly. In particular, in contrast to the previous studies that generally exploit a 1-D similarity vector based detector, a 2-D similarity matrix based decoder is specially designed to make full use of spatial information in the entire image, which better models the spatial contexts and leads to more accurate localization results. The modules are presented with details subsequently.}
	
	\begin{figure}[t]
		\centering
		\includegraphics[width=1.0\columnwidth]{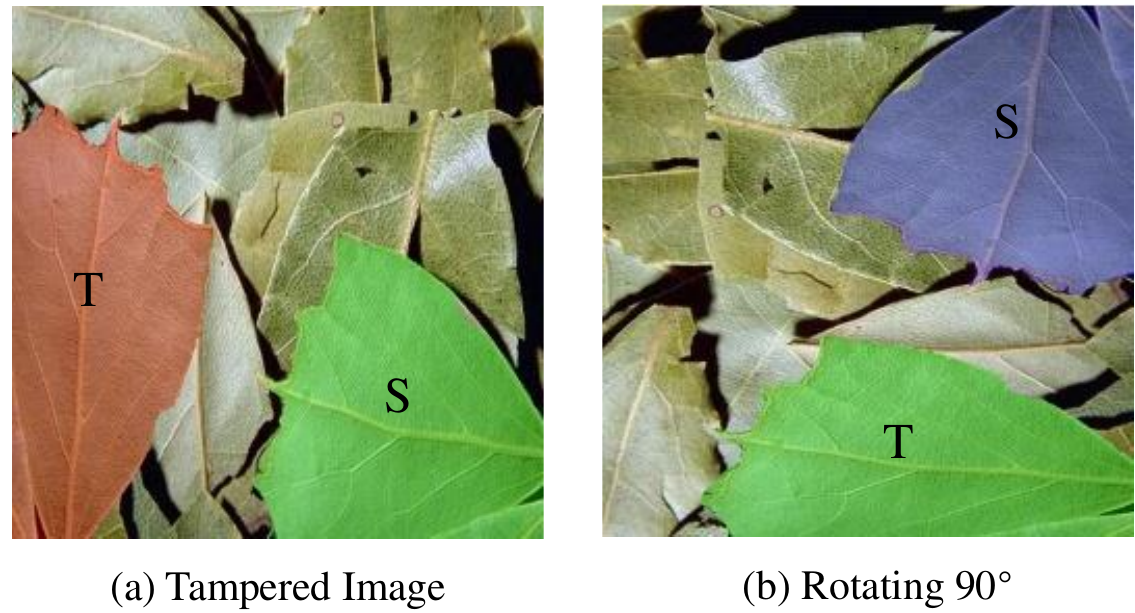}
		\caption{Illustration of challenges caused by rotating. \textbf{S} and \textbf{T} represent the source and target regions, respectively. The same color indicates a high visual similarity.}
		\label{pyramid_example}
	\end{figure}
	
	\subsection{Multi-directional Representation}

	The main challenge in tampered region detection is incurred by rotation variations between source and target regions and scale changes, and these issues are more intractable in CMFD than in the general object detection and recognition tasks. Fig. \ref{pyramid_example} displays one example, where the input image is tampered by rotating the source region by $90$ degrees and pasting it back into the target region of the image. As Fig. \ref{pyramid_example} (a) shows, the existing CNN-based approaches tend to fail due to their intrinsic limitation in modeling those changes, especially when the number and diversity of training samples are not adequate. 
	
	For rotation and scale robust representation in CMFD, we propose a multi-directional and multi-scale representation, which substantially augments the input image space with diverse orientations as well as sizes, making it easier to capture the source and target similarity. Specifically, given an image $\mathbf{I}$, after reshaping it into the fiducial size of $256\times 256$, a multi-directional image set is built by rotating it to four orientations (Fig. \ref{pyramid_example} (b) provides a simplified demonstration), \emph{i.e.} $\mathbf{I_{rot}} = \{\mathbf{I_\theta}\}, {\theta\in(0, \pi/2, \pi, 3\pi/2)}$. Similarly, a zoomed-in patch set with scale changes is constructed by enlarging the image twice and slicing out four non-overlapping patches, \emph{i.e.} $\mathbf{I_{sca}} = \{\mathbf{I_r}\}, {r\in (1, 2, 3, 4)}$.
	
	A number of CNNs with shared weights are then applied to the multi-directional image set and zoomed-in patch set for feature extraction, and here, we adopt the first 26 layers in a pre-trained VGG16-BN model, and the parameters are frozen during the training stage. In this way, the augmented feature sets $\mathbf{F_{aug}}$, including $\mathbf{F_{rot}}$ and $\mathbf{F_{sca}}$, are constructed from $\mathbf{I_{rot}}$ and $\mathbf{I_{sca}}$, respectively. By selecting basic feature $\mathbf{F_{base}} (\mathbf{F_{\theta=0}}$) from the original input image and one feature $\mathbf{F_{\theta}}$ from $\mathbf{F_{rot}}$, 4 pairs of features, $\mathbf{P = \{<F_{base}, F_{\theta}>\}}_{\theta\in(0, \pi/2, \pi, 3\pi/2)}$, are created. Each feature pair in $\mathbf{P}$ is further fed into the two-stream tampered region detector to localize the similar regions between $\mathbf{I_{\theta=0}}$ and $\mathbf{I_{\theta}}$. 
	Similarly, the other 4 feature pairs are developed from $\mathbf{F_{sca}}$, which mainly cope with scale changes. The predicted tampered regions of the patch set are resized and pasted into their original locations. Finally, the predicted mask is produced by taking the maximum value in each pixel from the 8 duplicated detectors.

	\begin{figure}[tp]
		\centering
		\includegraphics[width=1\columnwidth]{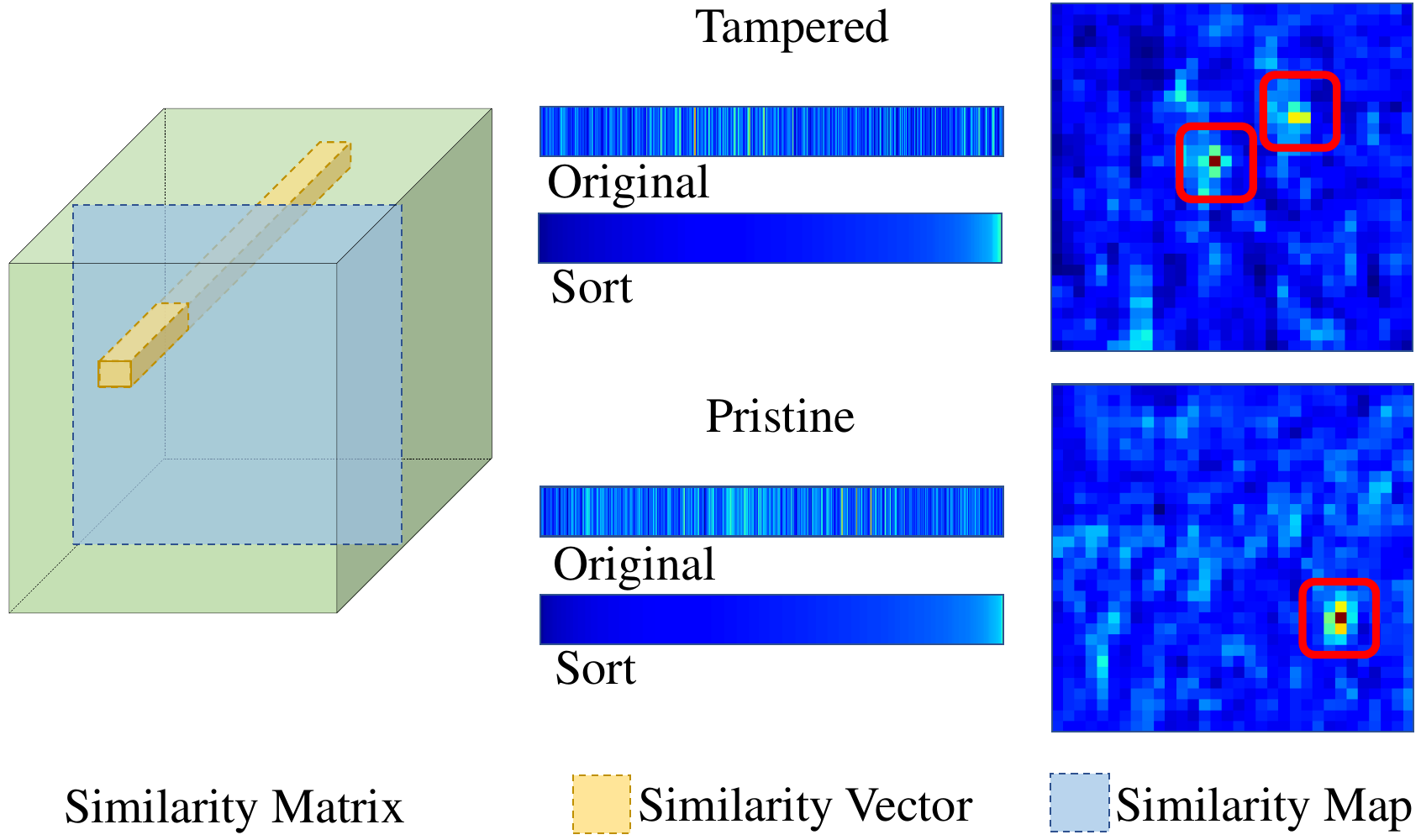}
		\caption{Comparison of similarity vector and similarity map from tampered and pristine regions. Hot color indicates high similarity.}
		\label{similarity map}
	\end{figure}
	
	\subsection{Tampered Region Detection}

As previously stated, to detect tampered region, the current deep models \cite{DBLP:conf/eccv/WuAN18,DBLP:journals/tifs/ZhongP20,DBLP:conf/cvpr/IslamLBH20} generally exploit a simple 1-D similarity vector based decoder, but which tend to drop crucial spatial information. To solve this problem, we design a novel decoding mechanism based on the 2D similarity matrix to make full use of raw similarities with original spatial clues. In particular, the proposed tampered region detector consists of two modules: the Similarity Computation Layer and the Similarity Map Classifier. 
	
	\textbf{Similarity Computation Layer}
	This module measures the similarities between feature pairs, where the source and target regions generally possess a higher similarity than the other regions. We calculate the visual similarity matrix as:
	\begin{equation}
	\emph{S$^{\prime}$} = (\emph{$\hat{F}_i$}\emph{$\hat{F}_j^{T}$}) / c, 
	\end{equation}
	where ($\hat{\cdot}$) represents the operation to normalize and flatten the input vectors from $\mathbf{F} \in{\mathbb{R} ^{32\times32\times512}}$ to $\mathbf{\hat{F}} \in{\mathbb{R} ^{1024\times512}}$ and $c$ denotes the normalized constant. We then reshape the similarity matrix $\mathbf{S^{\prime}}$ into a tensor $\mathbf{S} \in{\mathbb{R} ^{32\times32\times1024}}$. It is further processed by the Similarity Map Classifier.

	\textbf{Similarity Map Classifier}
     Treating $\mathbf{S} \in {\mathbb{R} ^{32\times32\times1024}}$ as $32\times32$ dimensional similarity vectors is a common practice in previous studies \cite{DBLP:conf/eccv/WuAN18,DBLP:journals/tifs/ZhongP20,DBLP:conf/cvpr/IslamLBH20}, \emph{i.e.} $\mathbf{S} = \{\mathbf{v}_{i, j}\}_{i,j\in{[0,...,31]}}$ where ${v}_{i, j}$ represents the similarities between the features from patch ($i$, $j$) and the others. 
     Some methods further sort the vectors and consider only the top percentage similarities values, \textit{e.g.}, BusterNet \cite{DBLP:conf/eccv/WuAN18}. However, such mechanism may discard values useful in determining the location results. Meanwhile, it does not well preserve the 2D spatial location information, dropping the significant context information from the surrounding patches. As shown in the 2D similarity map in Fig. \ref{similarity map}, the numbers of peaks (marked by red circles) in the pristine and tampered images are different. There is only one peak, which represents the similarity to itself, from the pristine regions, while there exists at least one more peak from the tampered regions revealing another highly similar area, and its surrounding patches also have relatively higher scores. This difference can be easily observed on the 2-D similarity maps. By contrast, as spatial clues are discarded either in the original or the sorted 1-D similarity vectors, they cannot identify such a difference, leading to the failure in tampered region localization.   
    
\begin{figure*}[ht]
	\centering
	\includegraphics[width=2.06\columnwidth]{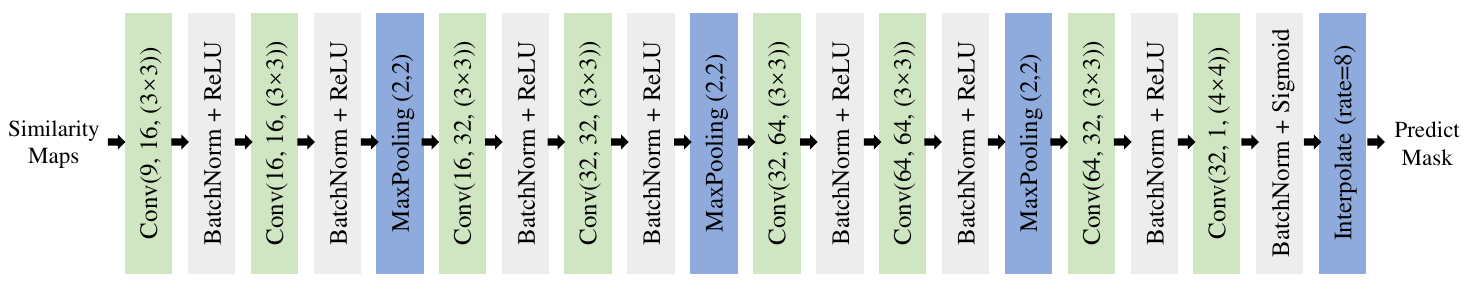}
	\vspace{0.01cm}
	\caption{Detailed architecture of the Similarity Map Classifier, which predicts the 2D tampered region mask.}
	\label{net}
	\end{figure*}

    To sufficiently leverage the spatial context of the image, we analyze $\mathbf{S}$ in the 2D manner and the task is transformed to classifying the similarity maps. Concretely, $\mathbf{S}$ is regarded as $1,024$ similarity maps with the size of ${32\times32}$, \emph{i.e.} $\mathbf{S} = \{\mathbf{m}_{j + i\times32}\}_{i, j\in{[0,...,31]}}$ where $\mathbf{m}_{j+i\times32} \in {\mathbb{R} ^{32\times32}}$ denotes the similarity map of patch ($i$, $j$). In this case, $\mathbf{m}$ can not only includes all the information in ${v}$, but also preserves the original spatial structures. 
    Based on the spatial clues above, we further perform spatial context aggregation and design an additional classifier to exploit the neighboring areas for robustness enhancement. To integrate the information around ${\mathbf{m}_{j+32\times i}}$ from patch ${p_{i,j}}$, we use the similarity maps from the eight patches which are spatially adjacent and concatenate them with $\mathbf{m}$, \emph{i.e.} \{\emph{p$_{i+m,j+n}$}\}$_{m,n\in{(-1,0,1)}}$, as the input of the similarity classifier. Theoretically, various CNN networks can serve as the similarity map classifier. In particular, we design our classifier consisting of eight convolution layers interspersed with three max pooling layers to generate the mask, as depicted in Fig. \ref{net}.
	
\section{Experiments}
\label{Experiments}

To validate the proposed CMFD approach, we carry out extensive experiments and make fair comparison to the state-of-the-art counterparts. The databases, implementation details, and experimental results are presented in the subsequent.

	\subsection{Databases}
    	In this study, we create the first copy-move forgery database with samples generated by three different neural networks, which is complementary to the current hand-crafted ones. We carry out comprehensive experiments on two major public benchmarks, \emph{i.e.} CASIA CMFD \cite{DBLP:conf/eccv/WuAN18} and CoMoFoD \cite{tralic2013comofod}, and the newly presented one. All the experiments are performed under a cross-dataset protocol.
    	
    	\textbf{CASIA CMFD and CoMoFoD.}  The CASIA CMFD dataset contains $2,626$ images, \emph{i.e.} $1,313$ tampered images and their original counterparts, which contain the attacks of rotation and scaling. The CoMoFoD dataset provides $5,000$ tampered images rendered from 200 base images with 25 manipulation categories, each of which corresponds to one type of attack. For more details, please refer to \cite{DBLP:conf/eccv/WuAN18}, \cite{tralic2013comofod}. 
    	
        \textbf{DCF: Deep-synthesized Copy-Move Forgery Database.}  DCF database provides 21,000 images generated by three deep manipulation approaches, corresponding to three subsets: DCF-VAE, DCF-Transfer, and GAN-Rewriting, respectively. DCF-VAE and DCF-Transfer subsets are achieved by post-processing the current manually tampered dataset (\textit{i.e.} CoMoFoD \cite{tralic2013comofod}) by neural networks, aiming to remove the clues or artifacts that used to be exploited as the foundation for detection and localization, \textit{e.g.}, edge and color inconsistencies. GAN-Rewriting subset exploits deep models to generate virtual images of diverse scenes and then manufacture forgery areas, allowing visually plausible results and better edge consistency around the tampered regions. The DCF database will be released to the community for future studies.
        
        1) \textit{DCF-VAE} set is created by applying the Variational Auto-Encoder (VAE) \cite{kingma2013auto} to reconstruct the tampered images in CoMoFoD \cite{tralic2013comofod}, which makes it have the same directory structure as CoMoFoD, consisting of 10,000 images of $512\times512$ pixels. The employed VAE model is implemented by a 20-layered encoder-decoder architecture (the detailed architecture can be found in \cite{kingma2013auto}), which is trained for 300 epochs in total by Adam optimizer, with a learning rate of 1e-4 and a batch size of 26. The MSE and KL scatter losses are exploited as usual, both with equal weighting factors. 
        
        2) \textit{DCF-Transfer} set is created by post-processing the manually tampered images in CoMoFoD by the pre-trained FastPhotoTransfer  \cite{li2018closed} model. The same samples are used as in DCF-VAE. Each image is generated by randomly selecting a style image that is not under attack from CoMoFoD and then transferring it to another manually tampered image. After style transfer, image smoothing is performed to prevent over-styling. In particular, we use a guided filter with $r=35$ and $eps=0.001$ to approximate standard smoothing and speed up the process. 
        
        3)  \textit{GAN-Rewriting} set provides 1,000 images that are wholly virtual. We first use StyleGAN-v2 \cite{Karras2019stylegan2} and Progressive GAN \cite{karras2017progressive} trained on the LSUN dataset to generate images and then manually select around 300 high-quality ones from different scenes, including church, bedroom, and kitchen. With the interactive interface of the Rewriting model \cite{bau2020rewriting}, an object in an image (usually larger than $20\times20$ pixels) is manually circled and moved, so that the corresponding features are copied and pasted in the feature maps to generate deep copy-move images. We adjust the target position to make the deep tampered image looks natural, and the tamper masks are recorded during this operation. We manipulate each image for 2 to 5 times with regions of different sizes manipulated, and finally 1,000 forged images are created along with the corresponding copy-move masks.

		\textbf{Manually Manipulated Training Data.} Since there is no available manually manipulated training data, we collect data from the MS COCO dataset \cite{lin2014microsoft}. Specially, given an image and the corresponding segmentation annotation, we traverse all the annotated objects within the image and select the objects with the area greater than 1,000 pixels and the minimum length of the bounding box greater than 64 pixels. We copy an object of suitable size according to the given mask, and then perform random rotation and scaling with the angles and rations ranging from [0\degree, 360\degree] and [0.8, 1.2], respectively. The transformed object is then randomly pasted to another location in the same image with equal probability. Besides, inspired by \cite{DBLP:conf/cvpr/TripathiCATRC19}, to avoid overfitting to object shapes, additional images are generated by randomly cropping regions from other images using object boundaries. Totally, we have $414,895$ manually manipulated images for training.

	\definecolor{my_color1}{gray}{0.6}
	
	\begin{table*}[t]
		\centering
		    \caption{Performance comparison of different methods in terms of precision, recall, and F1-score on CASIA CMFD \cite{DBLP:conf/eccv/WuAN18} and CoMoFoD \cite{tralic2013comofod}.}
		    \label{casia res}
		\begin{threeparttable}
			\begin{tabular}{p{3.0cm}|p{0.8cm}p{0.8cm}p{0.8cm}p{0.8cm}p{0.8cm}p{0.8cm}|p{0.8cm}p{0.8cm}p{0.8cm}p{0.8cm}p{0.8cm}p{0.8cm}}
				\toprule
				\multicolumn{1}{c}{} &\multicolumn{6}{c}{CASIA CMFD} & \multicolumn{6}{c}{CoMoFoD}\cr 
				\multirow{2}{*}{Methods}&
				\multicolumn{3}{c}{pixel-level}& \multicolumn{3}{c}{image-level}& \multicolumn{3}{c}{pixel-level}& \multicolumn{3}{c}{image-level}  \cr 
				\cmidrule(lr){2-4} \cmidrule(lr){5-7} \cmidrule(lr){8-10} \cmidrule{11-13}
				&\emph{Prec.}	&\emph{Rec.} 	&\emph{F1}  	&\emph{Prec.}	&\emph{Rec.} 	&\emph{F1} &\emph{Prec.}	&\emph{Rec.} 	&\emph{F1}  	&\emph{Prec.}	&\emph{Rec.} 	&\emph{F1}\cr
				\midrule
			    Block-ZM \textcolor{my_color1}{[IHW10]} & 32.89 & 28.34 & 29.26 & 74.65 & 45.09 & 56.22 & 8.09 & 8.33 & 7.28 & 52.94 & 23.18 & 31.36  \cr
                Ada-Seg \textcolor{my_color1}{[TIFS15]} & 47.71 & 13.16 & 18.78 & 88.04 & 52.70 & 65.94 & 38.49 & 16.89 & 21.09 & 75.87 & 51.08 & 60.80  \cr
                DenseField \textcolor{my_color1}{[TIFS15]}\tnote{*} & 20.55 & 20.91 & 20.36 & \textbf{99.51} & 30.61 & 46.82 & 22.23 & 23.63 & 22.60 & 80.34 & 20.10 & 32.15  \cr
                BusterNet \textcolor{my_color1}{[ECCV18]} & 55.71 & 43.81 & 45.51 & 76.47 & 70.30 & 73.25 & 56.50 & 46.97 & 47.80 & \textbf{80.82} & 71.38 & \textbf{75.74}  \cr
                DenseIncep \textcolor{my_color1}{[TIFS19]}\tnote{*} & 70.85 & 58.85 & 64.29 & - & - & - & 46.10 & 42.20 & 44.10 & - & - & -  \cr
                ManTraNet \textcolor{my_color1}{[CVPR19]} & 28.72 & 44.13 & 31.93 & 73.25 & 70.30 & 71.75 & 21.46 & 18.89 & 10.44 & 50.91 & 80.48 & 61.75  \cr
                DOA-GAN \textcolor{my_color1}{[CVPR20]} & 53.92 & 41.51 & 42.26 & 62.90 & 77.61 & 69.49 & 50.12 & 43.67 & 40.69 & 59.97 & 82.48 & 69.16  \cr
                AR-Net \textcolor{my_color1}{[TII20]}\tnote{*}  & 58.32 & 37.33 & 45.52 & - & - & - & 54.21 & 46.55 & 50.09 & - & - & -  \cr
                CMSDNet \textcolor{my_color1}{[TMM20]}  & 52.51 & 52.48 & 48.64 & 56.02 & 88.88 & 68.73 & 48.08 & 57.01 & 48.49 & 53.07 & 92.88 & 67.53  \cr
                SuperGlue \textcolor{my_color1}{[TIP21]}\tnote{*}   & 64.94 & 45.20 & 47.82 & 78.60 & 76.09 & 77.32 & - & - & - & - & - & -  \cr
                \textbf{Ours} & \textbf{76.77} & \textbf{73.10} & \textbf{72.09} & 67.72 & \textbf{94.44} & \textbf{78.88} & \textbf{57.64} & \textbf{65.86} & \textbf{56.80} & 58.72 & \textbf{96.00} & 72.87  \cr
				\bottomrule
			\end{tabular}

			\begin{tablenotes}
                \footnotesize       
                \item[*] We quote the results from the original papers since the codes are not available.
            \end{tablenotes} 
		\end{threeparttable}
	\end{table*}

      \definecolor{my_color1}{gray}{0.6}
    \begin{table*}[!ht]
    \centering
         \caption{Performance comparison of methods in terms of precision, recall, and F1-score on DCF-VAE and DCF-Transfer subsets.}
        \label{post-processed res}
    \begin{threeparttable}
            \begin{tabular}{p{3.0cm}|p{0.8cm}p{0.8cm}p{0.8cm}p{0.8cm}p{0.8cm}p{0.8cm}|p{0.8cm}p{0.8cm}p{0.8cm}p{0.8cm}p{0.8cm}p{0.8cm}
            }
                    \toprule
                    \multicolumn{1}{c}{} &\multicolumn{6}{c}{DCF-VAE}
                    & \multicolumn{6}{c}{DCF-Transfer}  
                    \cr \multirow{2}{*}{Methods}&
                    \multicolumn{3}{c}{pixel-level}& \multicolumn{3}{c}{image-level}& 
                    \multicolumn{3}{c}{pixel-level}& \multicolumn{3}{c}{image-level} \cr 
                    \cmidrule(lr){2-4} \cmidrule(lr){5-7} 
                    \cmidrule(lr){8-10} \cmidrule{11-13} 
                    &\emph{Prec.}    &\emph{Rec.}     &\emph{F1}      &\emph{Prec.}    &\emph{Rec.}     &\emph{F1}
                    &\emph{Prec.}    &\emph{Rec.}     &\emph{F1}      &\emph{Prec.}    &\emph{Rec.}     &\emph{F1}\cr
                    \midrule
                    Block-ZM \textcolor{my_color1}{[IHW10] }& 0.00 & 0.00 & 0.00 & 8.00 & 0.04 & 0.08 & 0.00 & 0.01 & 0.00 & 48.33 & 0.60 & 1.17 \cr
    Ada-Seg \textcolor{my_color1}{[TIFS15]} & 6.32 & 2.93 & 2.76 & 57.08 & 17.92 & 27.08 & 6.71 & 2.25 & 2.86 & 62.29 & 15.16 & 24.03 \cr
    BusterNet \textcolor{my_color1}{[ECCV18]} & 42.21 & 32.84 & 33.10 & \textbf{65.55} & 66.62 & 65.92 & 49.13 & 38.24 & 39.14 & \textbf{69.07} & 69.08 & 69.01 \cr
    ManTraNet \textcolor{my_color1}{[CVPR19]} & 17.00 & 14.27 & 10.72 & 49.61 & 65.40 & 55.99 & 18.31 & 8.72 & 7.92 & 50.35 & 80.64 & 61.71 \cr
    CMSDNet \textcolor{my_color1}{[TMM20] }& 38.11 & 36.62 & 31.88 & 52.89 & 91.12 & 66.90 & 42.57 & 49.75 & 42.27 & 53.56 & 88.58 & 66.74  \cr
    DOA-GAN \textcolor{my_color1}{[CVPR20]} & 36.61 & 31.96 & 28.03 & 54.10 & 83.64 & 65.54 & 46.50 & 43.61 & 39.70 & 57.48 & 86.60 & 69.01  \cr
    \textbf{Ours} & \textbf{49.98} & \textbf{37.75} & \textbf{37.81} & 59.93 & 87.50 & \textbf{71.14} & \textbf{56.17} & \textbf{52.91} & \textbf{49.33} & 62.37 & \textbf{92.00} & \textbf{74.34}  \cr
    \textbf{Ours\_Finetune\tnote{*}} & \textbf{49.51} & \textbf{49.26} & \textbf{41.29} & 58.26 & \textbf{93.50} & \textbf{71.79} & \textbf{49.81} & \textbf{65.68} & \textbf{50.39} & 57.27 & \textbf{94.50} & \textbf{71.32} \cr
                    \bottomrule
            \end{tabular}
       
			\begin{tablenotes}
                \footnotesize       
                \item[*] Around 30\% and 10\% of the training data is post-processed by VAE \cite{kingma2013auto} and style transfer \cite{li2018closed}. The deep synthesized images replace the original ones to fine-tune the network.
            \end{tablenotes}    
    \end{threeparttable}
    \end{table*}

    \begin{figure*}[ht]
	    \centering
		\subfloat{\includegraphics[width=0.33\textwidth]{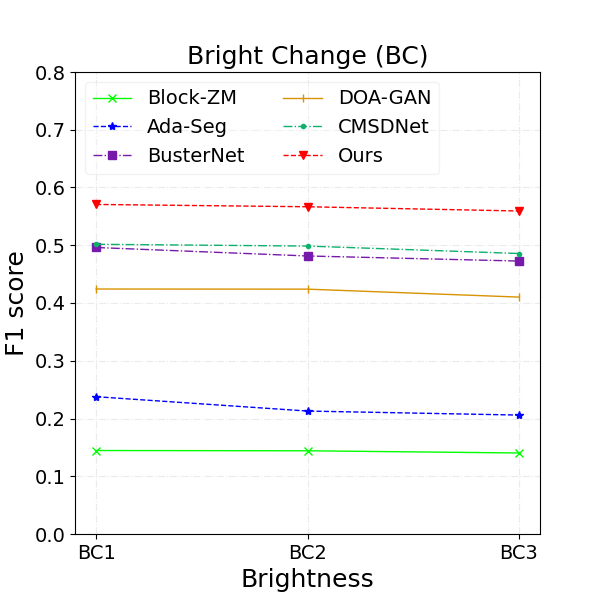}}
		\hfill
		\subfloat{\includegraphics[width=0.33\textwidth]{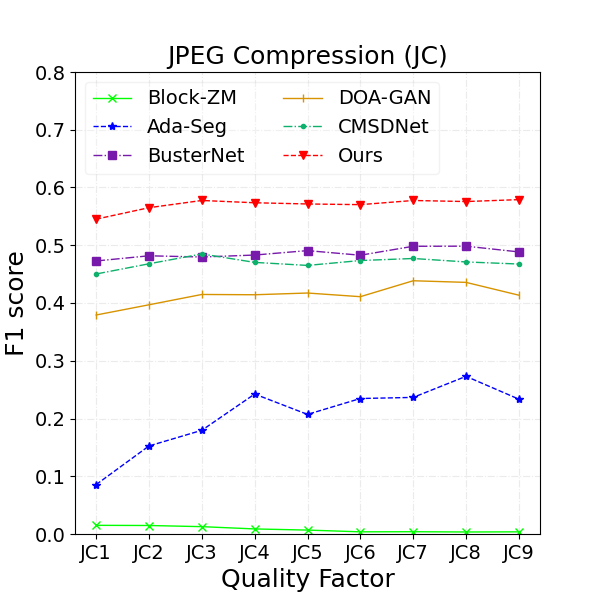}}
		\hfill
		\subfloat{\includegraphics[width=0.33\textwidth]{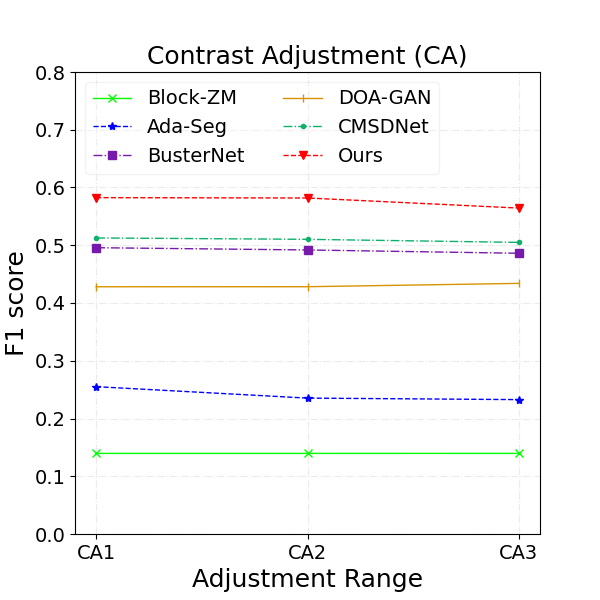}}
		\newline

		\subfloat{\includegraphics[width=0.33\textwidth]{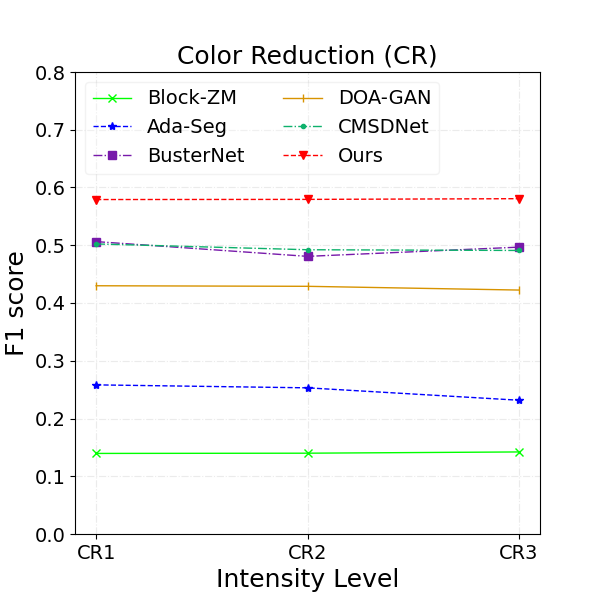}}
		\hfill
		\subfloat{\includegraphics[width=0.33\textwidth]{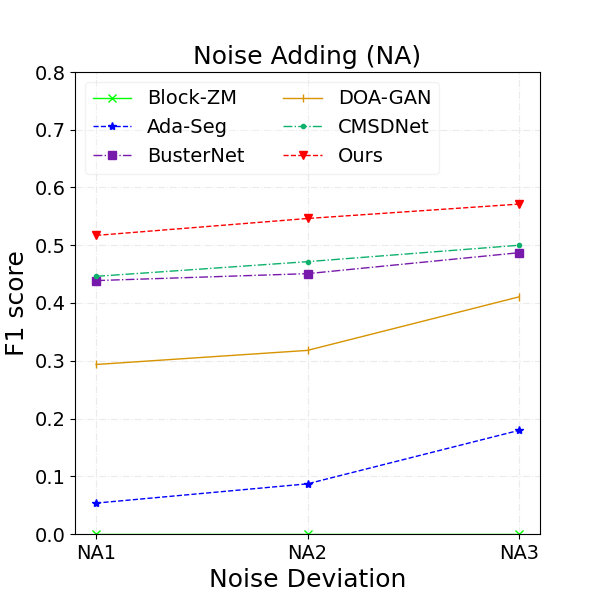}}
		\hfill
		\subfloat{\includegraphics[width=0.33\textwidth]{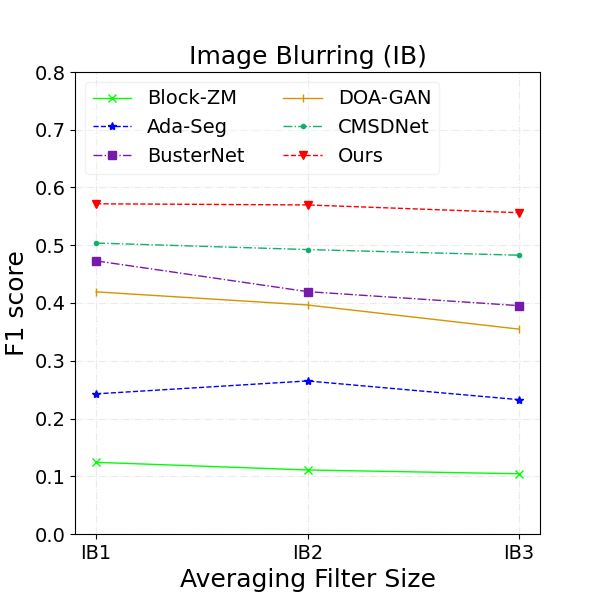}}
	
		\caption{Pixel-level F1-score on the CoMoFoD dataset \cite{tralic2013comofod} under different levels of attacks. (a) Bright changes; (b) JPEG compression; (c) contrast adjustment; (d) color reduction; (e) noise adding; and (f) image blurring.}
		\label{fig comofod res}
	\end{figure*}
	
	\begin{figure*}[!ht]
		\centering
		\includegraphics[width=2.05\columnwidth]{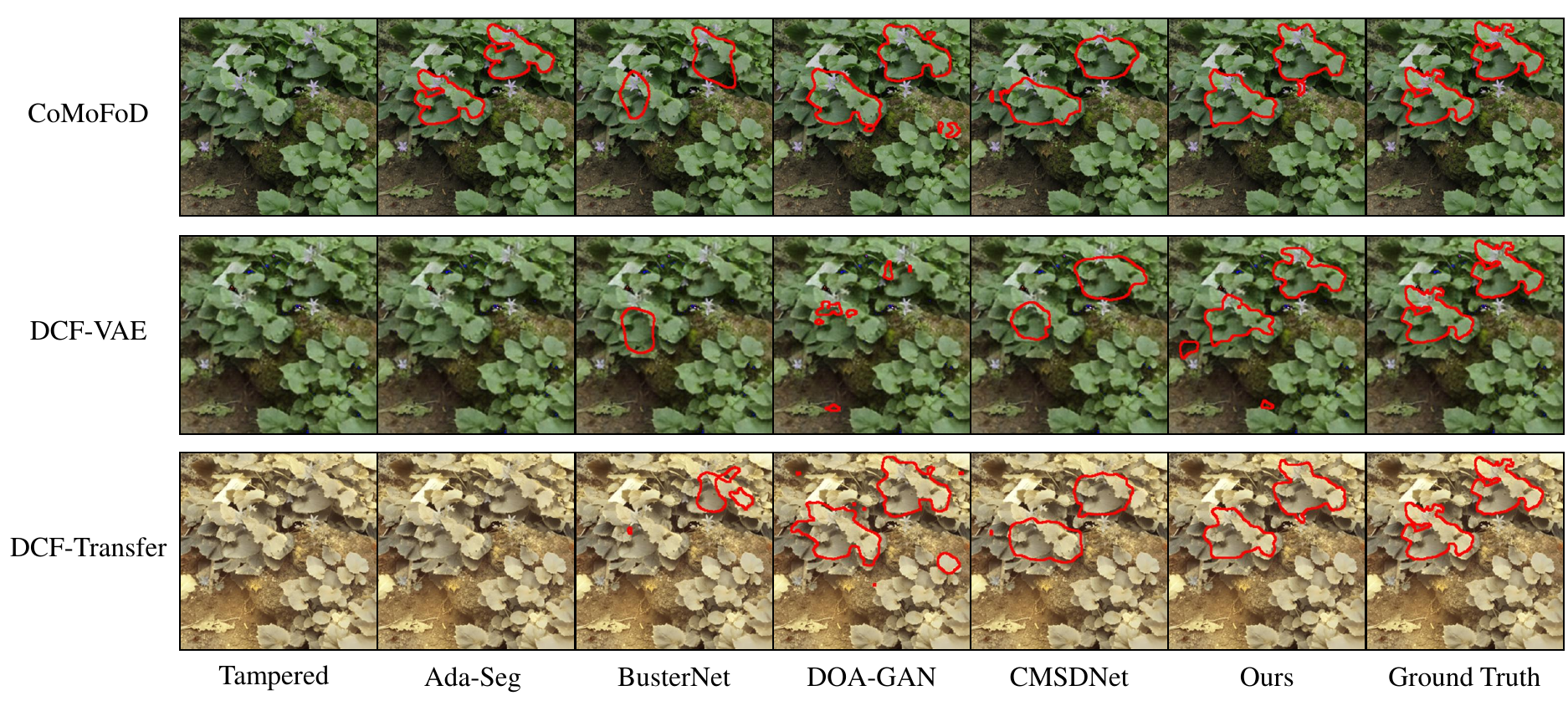}
		\caption{Visualization comparison on the images from the CoMoFoD dataset \cite{tralic2013comofod}, DCF-VAE, and DCF-Transfer. From top to bottom are the tampered images in CoMoFoD, the post-processed images using Variational Auto-Encoder (VAE) \cite{kingma2013auto} and style transfer \cite{li2018closed}. From left to right are tampered images; results of Ada-Seg \cite{DBLP:journals/tifs/PunYB15}, BusterNet \cite{DBLP:conf/eccv/WuAN18}, DOA-GAN \cite{DBLP:conf/cvpr/IslamLBH20}, CMSDNet \cite{chen2020serial}, and our method; and ground truth masks.}
		\label{visual_edge}
	\end{figure*}
		
	\subsection{Implementation Details}
    	In our detection network, the feature extractor is initialized with the VGG16-BN model pre-trained using ImageNet \cite{DBLP:conf/cvpr/DengDSLL009} and the other modules with the default setting in PyTorch. Training is on $4\times$ NVIDIA V100 GPUs. The Adam algorithm is used for optimization, with a batch size of $64$, a momentum of $0.9$, and a weight decay of $0.0005$. The initial learning rate and gamma are set to $1e^{-4}$ and $0.1$, and the learning rate is reduced by half every 20 epochs. The pixel-level binary cross-entropy is adopted as the loss function. 
    	
        At the training stage, for each predefined direction ${\theta\in(0, \pi/2, \pi, 3\pi/2)}$, we only rotate the images that the angle difference between the source and target objects is within [$\theta$-45\degree, $\theta$+45\degree]. In this way, around 130,000 samples are obtained for each augmented direction, and four detectors are trained accordingly. Meanwhile, the detector of 0\degree~is also trained with the augmented data with scaling changes.

	\subsection{Evaluation Protocols}
	
	We use the precison (Prec), recall (Rec), and F1-score (F1) as metrics in both the pixel-level and the image-level. For pixel-level evaluation, we compute precision, recall and F1-score for each image and report their average values. Only forged images are taken into account because F1-score is ill-defined when there is no positive pixel. For image-level evaluation, a query image is judged as positive when more than 0.2\% pixels are predicted as tampered. To filter out some false alarms of predicted masks, the results with the detection area smaller than 0.1\% of the image size are excluded. We compare the proposed method with a number of existing studies, including both traditional CMFD methods, \emph{i.e.}, Block-ZM \cite{DBLP:conf/ih/RyuLL10}, Ada-Seg \cite{DBLP:journals/tifs/PunYB15}, and DenseField \cite{DBLP:journals/tifs/CozzolinoPV15}, and the deep learning based ones, \emph{i.e.}, BusterNet \cite{DBLP:conf/eccv/WuAN18}, DenseIncep \cite{DBLP:journals/tifs/ZhongP20}, MantraNet \cite{DBLP:conf/cvpr/0001AN19}, DOA-GAN \cite{DBLP:conf/cvpr/IslamLBH20}, AR-Net \cite{zhu2020ar}, CMSDNet \cite{chen2020serial}, and SuperGlue \cite{liu2021two}. 
	
	To make fair comparison with the state-of-the-art counterparts, all evaluations are conducted with the original image resolution or the optimal resolution of the models. For the models without available codes, \emph{i.e.}, DenseField \cite{DBLP:journals/tifs/CozzolinoPV15}, DenseIncep \cite{DBLP:journals/tifs/ZhongP20} and AR-Net \cite{zhu2020ar} and SuperGlue \cite{liu2021two}, we report the results in the papers directly. In particular, for SuperGlue \cite{liu2021two}, we only report the results on CASIA CMFD due to the incomplete results on CoMoFoD.

	\subsection{Quantitative Experimental Results}
	
	\subsubsection{CASIA CMFD} 
	
	The left part of Table \ref{casia res} shows the results of different methods on the CASIA CMFD benchmark. As we can see in the table, our approach outperforms the others in terms of F1-score in both the pixel-level and the image-level. Moreover, it also reaches a higher recall compared with the other baselines, which demonstrates that our model locates more tampered regions. Such a superiority is achieved in two-fold: (1) based on the multiple-directional architecture and augmentation in both rotations and scales, the method delivers a more powerful representation of the sampled regions, which well balances the distinctiveness between similar objects and the robustness of the same object under visual variations; (2) the similarity map classifier preserves crucial spatial information to discriminate tampered regions, which further boosts the performance.
	
	\subsubsection{CoMoFoD} To validate the robustness of our method to various attacks, we carry out experiments on the CoMoFoD dataset and report the overall performance in Table \ref{casia res}. As shown in the right part of the table, similar to the cases on CASIA CMFD, our approach consistently surpasses the other counterparts. This evaluation involves 25 categories of attacks, including JPEG compression (JC), noise adding (NA), image blurring (IB), brightness changing (BC), color reduction (CR), and contrast adjustment (CA), each with a different level represented by a number behind. We report detailed scores in each category, and as displayed in Fig. \ref{fig comofod res}, our approach works in a very stable way, under different levels of attacks. Since there are limited scaling diversity and rotation variations on the samples, the results suggest that the proposed similarity map classifier contributes to this robustness enhancement.
	
	\definecolor{my_color1}{gray}{0.6}
    \begin{table}[!t]
    \centering
      \caption{Performance comparison of different methods in terms of precision, recall, and F1-score in the pixel-level.
      }
      \label{rewriting res}
    \begin{threeparttable}
        \begin{tabular}{p{3.6cm}|C{1.1cm}C{1.1cm}C{1.1cm}}
                \toprule
                (a) GAN-Rewriting &\emph{Prec.}    &\emph{Rec.}     &\emph{F1}\cr
                \midrule
                Block-ZM \textcolor{my_color1}{[IHW10] }& 1.25 & 0.28 & 0.45 \cr
                Ada-Seg \textcolor{my_color1}{[TIFS15]} & 7.72 & 1.10 & 1.80 \cr
                BusterNet \textcolor{my_color1}{[ECCV18]} & 53.50 & 25.76 & 31.61 \cr
                ManTraNet \textcolor{my_color1}{[CVPR19]} & 41.53 & 17.21 & 20.19 \cr
                CMSDNet \textcolor{my_color1}{[TMM20] } & 60.69 & \textbf{41.90} & 45.84  \cr
                DOA-GAN \textcolor{my_color1}{[CVPR20]} & 53.71 & 17.25 & 23.11  \cr
                \textbf{Ours} & \textbf{72.73} & 40.22 & \textbf{49.10}  \cr
                \toprule
                (b) GAN-CopyMove &\emph{Prec.}    &\emph{Rec.}     &\emph{F1}\cr
                
                \midrule
                Block-ZM \textcolor{my_color1}{[IHW10] }& 74.18   & 65.89   & 69.30 \cr 
                Ada-Seg \textcolor{my_color1}{[TIFS15]} & 16.70   & 3.36    & 5.23 \cr 
                BusterNet \textcolor{my_color1}{[ECCV18]} & 69.26   & 42.45   & 49.03 \cr
                ManTraNet \textcolor{my_color1}{[CVPR19]} & 40.42   & 15.89   & 18.81 \cr 
                CMSDNet \textcolor{my_color1}{[TMM20] } & 78.22   & 50.21   & 56.66 \cr 
                DOA-GAN \textcolor{my_color1}{[CVPR20]} & 68.00   & 55.45   & 57.31 \cr 
                \textbf{Ours} & \textbf{88.45}   & \textbf{86.40}   & \textbf{86.54} \cr 
                \bottomrule
        \end{tabular}

    \end{threeparttable}
    \end{table}

	\begin{figure*}[!h]
		\centering
		\includegraphics[width=2.05\columnwidth]{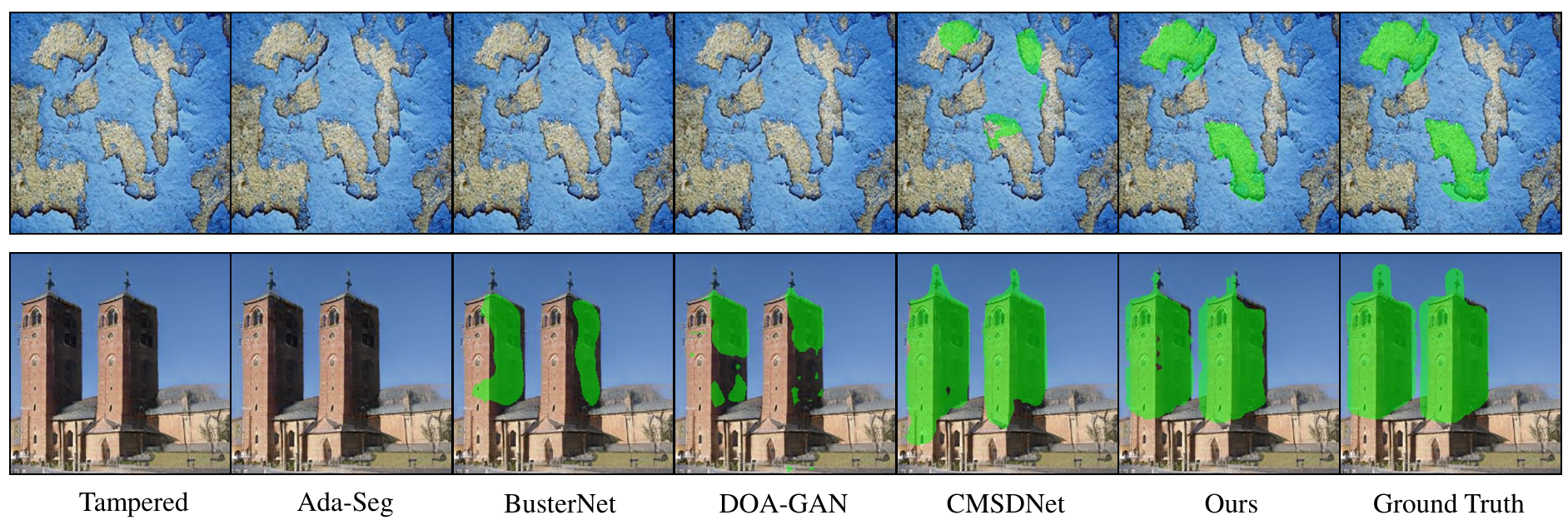}
		\caption{Visualization comparison on the CASIA CMFD \cite{DBLP:conf/eccv/WuAN18} and GAN-Rewriting datasets. From left to right are the tampered images; results of Ada-Seg \cite{DBLP:journals/tifs/PunYB15}, BusterNet \cite{DBLP:conf/eccv/WuAN18}, DOA-GAN \cite{DBLP:conf/cvpr/IslamLBH20}, CMSDNet \cite{chen2020serial}, and our method; and ground truth masks.}
		\label{visual}
	\end{figure*}

		\begin{figure*}[t]
		\centering
		\includegraphics[width=2.05\columnwidth]{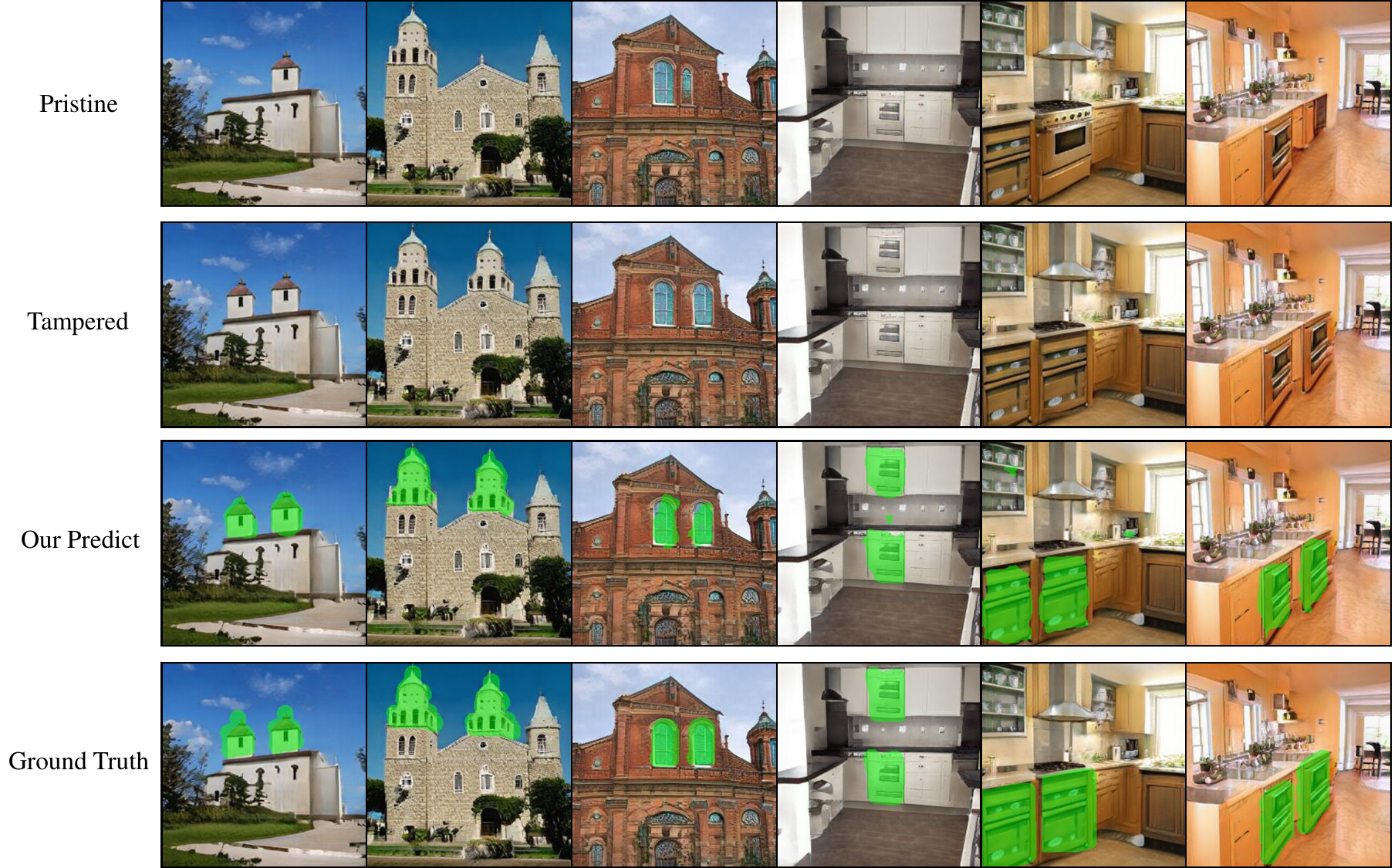}
		\vspace{0.01cm}
		\caption{Visualization results on the GAN-Rewriting subset. From top to bottom are the pristine synthetic images, the images tampered by deep rewriting method \cite{bau2020rewriting}, CMFD results of our model, and the ground truth masks.}
		\label{gan-rewriting}
	\end{figure*}

	\subsubsection{DCF Dataset\label{chp:synthetic}} To investigate the generalization performance on deep synthetic data, evaluations are further conducted on the DCF database. 
	
	\textit{DCF-VAE set}. As can be seen from the left part of Table \ref{post-processed res}, all the compared methods do not behave as well as on the manually manipulated data, especially the traditional approaches, \emph{i.e.}, Block-ZM \cite{DBLP:conf/ih/RyuLL10} and Ada-Seg \cite{DBLP:journals/tifs/PunYB15}, which exploit hand-crafted feature based analysis, are almost completely ineffective for such tampering type. The reason mainly lies in that the deep-synthesized images no longer convey clues that used to be exploited as the evidence for detection and localization, \textit{e.g.,} high-frequency artifacts around the copy-move edges, making the hand-crafted methods problematic in this situation. The results above motivate us to pay more attention to texture similarity for more accurate deep-synthesized copy-move detection. Among the compared deep learning-based approaches, the proposed method (the penultimate row in Table \ref{post-processed res}) outperforms the other counterparts in terms of F1-score in both pixel-level (37.81\%) and image-level (71.14\%), which clearly validates its effectiveness in texture similarity matching.

	\textit{DCF-Transfer set.} This subset is achieved by post-processing CoMoFoD. The results achieved on DCF-Transfer set are shown in the right part of Table \ref{post-processed res}. In this evaluation, the operation of style transfer causes a domain gap between the training set and the test set, which leads to performance degradation of all the compared CMFD methods. Nevertheless, the proposed approach surpasses all its counterparts with F1-score as the evaluation metric.

	In addition to the evaluations above, we further incorporate the deep synthetic data into the training set to check the performance changes of our approach on both DCF-VAE and DCF-Transfer subsets. In particular, around 30\% and 10\% of the original manually manipulated training data is post-processed by VAE \cite{kingma2013auto} reconstruction and style transfer \cite{li2018closed}. We substitute them with the deep synthesized ones and fine-tune the detection model. As can be seen from the last row in Table \ref{post-processed res}, the fine-tuning improves F1-score by 3.48\% and 1.06\% in pixel-level on DCF-VAE and DCF-Transfer, respectively. A slight performance gain is achieved by our method while the overall performance remains relatively stable. The results indicate that our method tends to have a good generalization performance to disturbance.

	\textit{GAN-Rewriting set.} In order to explicitly compare the performance difference of the CMFD methods on hand-crafted and deep-synthesized data, this evaluation is performed on the same source data but with different copy-move operations. In particular, during the generation process of \textit{GAN-Rewriting set}, we apply both GANs rewriting techniques \cite{bau2020rewriting} and the traditional copy-move operations on the intermediate images generated by StyleGAN2  \cite{Karras2019stylegan2} and Progressive GAN \cite{karras2017progressive}. In this way, a compared image set, \textit{i.e.,} GAN-CopyMove, is created. The results are demonstrated in the Table \ref{rewriting res}. We can observe that: 1) On the deep-synthesized rewriting data, our approach outperforms the others by a large margin in terms of precision and F1-score, clearly validating its effectiveness and generalizibilty, that it is less sensitive to the copy-move manners; 2) The tampering method can cause drastic changes in the detection results (11\%-68\% degradations for different methods in terms of F1-score), which informs us to be more cautious about deep-synthesized copy-move forgeries in further studies, and focusing on the texture semantic similarity tends to bring higher accuracy.

	\subsection{Visualization\label{chp:visualization}} 
	Figure \ref{visual_edge} shows the results on the images with the same content but from different databases, \textit{i.e.,} CoMoFoD \cite{tralic2013comofod}, DCF-VAE, and DCF-Transfer, respectively. The visualization demonstrates that the existing methods tend to detect unreasonable edges, since the clues such as edge inconsistency is no longer conveyed in the deep-synthesized images, whereas proposed method learns more accurate information for coping with this situation. 
	Figure \ref{visual} visualizes several example results obtained on the CASIA CMFD \cite{DBLP:conf/eccv/WuAN18} and GAN-Rewriting benchmarks, respectively. The first row shows that the proposed multi-directional similarity network learns more discriminative features for copy-move forgery region detection, while the second row indicates that the proposed method delivers more accurate masks than the other methods do. As can be seen, a large proportion of tampered pixels are neglected by the compared counterparts, while they can be correctly located by our method. In Fig. \ref{gan-rewriting}, we display more forgery data in the GAN-Rewriting set with the corresponding results achieved by the MSN model to further demonstrate its CMFD performance.

\begin{figure}[!t]
		\centering
		\includegraphics[width=1\columnwidth]{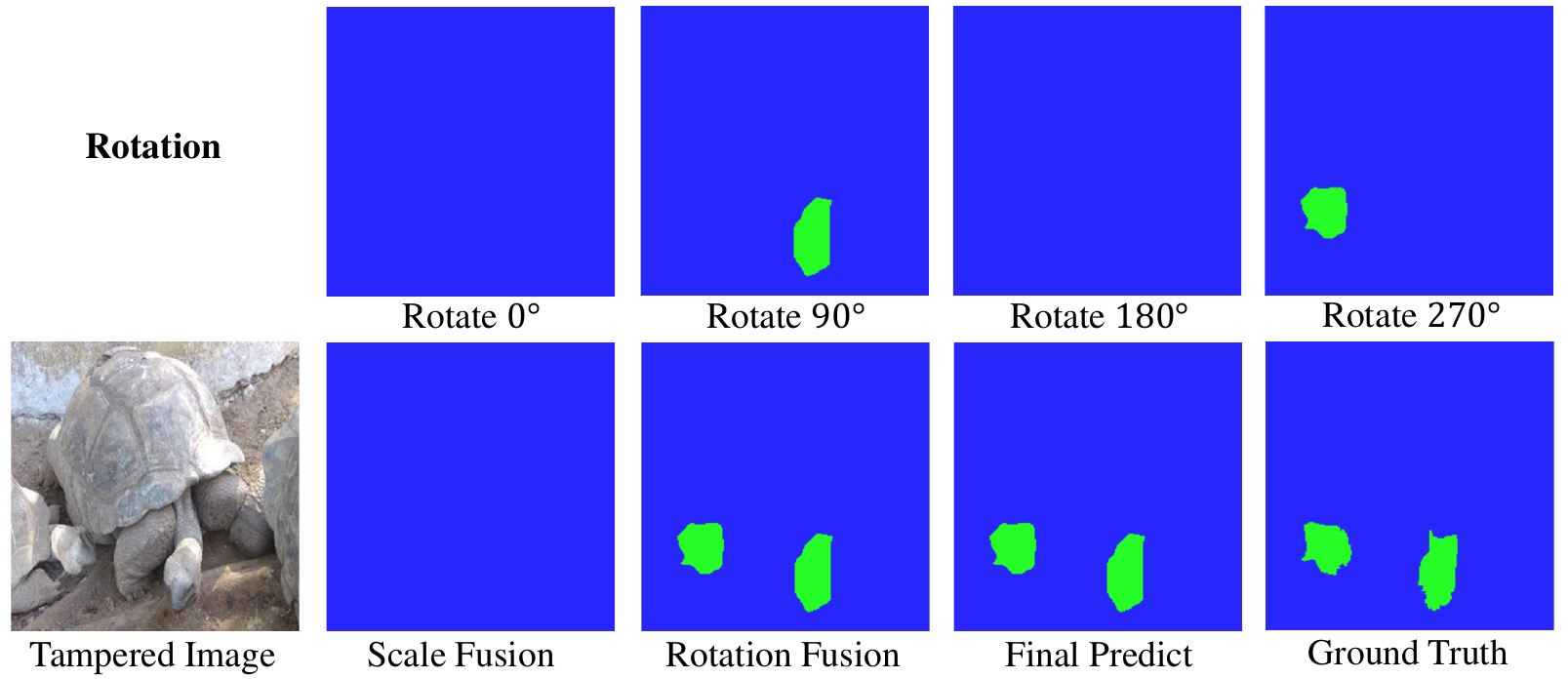}
		\vspace{0.01cm}
		\caption{Visualization of the results from the detectors.}
		\label{twelve_res}
	\end{figure}

	In order to have a better understanding of the impact of input space augmentation and multiple detectors, figure \ref{twelve_res} demonstrates the mask candidates predicted by our network. The input image is tampered by copying, rotating, and pasting the source region. The masks in the first row represent the outputs of detectors of four directions. We can observe that the tampered region detectors find that the source region from the image rotated 90 degrees is similar to the target region from the original image, and the target region from the image rotated 270 degrees is similar to the source region from the original image. Therefore, the second and fourth masks in the middle row reveal that the tampered candidates and the predicted mask successfully localize the entire tampered regions.
	
	\begin{table}
    \centering
    \begin{threeparttable}
        \caption{Ablation Studies on the CASIA CMFD dataset. 
        We re-implement the decoder of BusterNet \protect\cite{DBLP:conf/eccv/WuAN18} as our baseline, with its results shown in the first row.}
        \label{abl_res}
        \begin{tabular}{C{1.1cm}C{1.1cm}|C{1.1cm}|p{1cm}p{1cm}p{1cm}}
                \toprule
                \multicolumn{2}{c}{Augmentation} &\multicolumn{1}{c}{Detector} &\multicolumn{3}{c}{Pixel-level} \cr
                Rotation & Scaling & SMC & \emph{Prec.} & \emph{Rec.} & \emph{F1} \cr
                \midrule
                &&                                 &66.50 &69.02 &64.65\cr
                \checkmark&          &             &68.14 &69.08 &65.89\cr
                \checkmark&\checkmark&             &67.18 &72.72 &66.73\cr
                \checkmark&\checkmark&\checkmark   &\textbf{76.77}&\textbf{73.10}&\textbf{72.09}\cr
                \bottomrule
        \end{tabular}
    \end{threeparttable}
    \end{table}

	\subsection{Ablation Study}
	To validate the contributions of different components of the proposed method, we further perform ablation studies on the CASIA CMFD dataset. Here, we utilize a tampered region detector with the similarity decoder proposed in \cite{DBLP:conf/eccv/WuAN18} as our baseline.

	\textbf{Augmentation.} We first investigate the credits of rotation and scaling augmentation by integrating this augmented representation with the baseline tampered region detector. As shown in the 2nd row in Table \ref{abl_res}, the multi-directional network can better detect the forged images where the tampered regions undergo rotation attacks and hence increases the F1-score. Specifically, it improves the precision, recall, and F1-score by 1.64\%, 0.06\%, and 1.24\%, respectively. When incorporating scaling augmentation, the results continue to improve, which highlights the significance of such representation.
	
	\textbf{2D Similarity Map Classifier.} We further integrate the proposed similarity map classifier into the model. This tampered region detector improves the precision, recall, and F1-score by 9.59\%, 0.38\%, and 5.36\%, respectively, clearly validating the  effectiveness of proposed SMC decoder, which successfully leverages the spatial contexts to boost the CMDF performance.

    
	\subsection{Running Time}

	We record the inference time of our network on an NVIDIA 2080Ti and the total time to detect an image is ${112 ms}$ when the input resolution is resized to $256\times256$. It indeed inherits the property of fast processing of deep learning models. For comparison, Ada-seg \cite{DBLP:journals/tifs/PunYB15}, one of the fastest traditional methods, requires ${1.4 s}$ to process an image of the original higher resolution in the CASIA CMFD benchmark (corresponding to the results in Table \ref{casia res}) on a desktop equipped with the Core-i7 CPU and 8-GB RAM. Therefore, our model is faster than the traditional methods by an order of magnitude, which demonstrates its efficiency.

\section{Conclusion}
\label{conclusion}

	In this paper, we introduce an accurate and efficient multi-directional network for copy-move image forgery detection, along with the feature augmentation in scales and rotations to explicitly strengthen the representation of input images, which in particular improves the comparison accuracy between sampled regions under different scale and orientation variations. The tampered region detector with the 2-D similarity matrix based decoder is specifically designed to distinguish the tampered regions in a more powerful way, as the very informative spatial contexts are well utilized. Furthermore, to investigate the detection performance on deep synthetic data, we create the first copy-move forgery benchmark with samples generated by diverse neural networks, which enables more comprehensive evaluations on diverse deep copy-move forgeries. Extensive experiments are conducted on CASIA CMFD, CoMoFoD datasets and the newly presented benchmark, and state-of-the-art results are reported, clearly validating the effectiveness of the proposed method. Additionally, We provide the first experimental demonstration that almost all current detection methods encounter performance degradation of varying degrees on deep forgery data, which is a significant fact in the era of deep learning, and it provides helpful hints for the following studies on CMFD.
	
\newpage
\bibliographystyle{IEEEtran}
\bibliography{reference}
\end{document}